\documentclass[a4paper,twoside]{article}
\usepackage{float}

\usepackage{graphicx}
\usepackage{hyperref}
\usepackage{subcaption}
\usepackage{calc}
\usepackage{amssymb}
\usepackage{amstext}
\usepackage{amsmath}
\usepackage{amsthm}
\usepackage{multicol}
\usepackage{graphicx}
\usepackage{hyperref}
\usepackage{pslatex}
\usepackage{apalike}
\usepackage{algorithm2e}
\usepackage[bottom]{footmisc}
\usepackage{SCITEPRESS}     

\usepackage{fancyhdr}
\usepackage{lipsum} 
\fancypagestyle{plain}{
  \fancyfoot[L]{\tiny
    Belfarsi, E. A., Brubaker, S., and Valero, M.\\
    \textit{Optimizing Blood Transfusions and Predicting Shortages in Resource-Constrained Areas}.\\
    DOI: \href{https://doi.org/10.5220/0013182700003911}{10.5220/0013182700003911}\\
    Paper published under CC license (CC BY-NC-ND 4.0)\\
    In \textit{Proceedings of the 18th International Joint Conference on Biomedical Engineering Systems and Technologies (BIOSTEC 2025)} -- \textit{Volume 2: HEALTHINF}, pages 149–160\\
    ISBN: 978-989-758-731-3; ISSN: 2184-4305\\
    Proceedings Copyright \textcopyright~2025 by SCITEPRESS -- Science and Technology Publications, Lda.
  }
}

\begin{document}

\title{Optimizing Blood Transfusions and Predicting Shortages in Resource-Constrained Areas}
\author{\authorname{El Arbi Belfarsi\orcidAuthor{0009-0006-4145-5492}, Sophie Brubaker\orcidAuthor{0009-0000-3708-9000} Maria Valero\orcidAuthor{0000-0001-8913-9604}}
\affiliation{Kennesaw State University, 1100 South Marietta Pkwy SE, Marietta, GA 30060}
\email{ebelfars@students.kennesaw.edu, 
sophie.brubaker@wheelermagnet.com,
mvalero2@kennesaw.edu}
}

\abstract{Our research addresses the critical challenge of managing blood transfusions and optimizing allocation in resource-constrained regions. We present heuristic matching algorithms for donor-patient and blood bank selection, alongside machine learning methods to analyze blood transfusion acceptance data and predict potential shortages. We developed simulations to optimize blood bank operations, progressing from random allocation to a system incorporating proximity-based selection, blood type compatibility, expiration prioritization, and rarity scores. Moving from blind matching to a heuristic-based approach yielded a 28.6\% marginal improvement in blood request acceptance, while a multi-level heuristic matching resulted in a 47.6\% improvement. For shortage prediction, we compared Long Short-Term Memory (LSTM) networks, Linear Regression, and AutoRegressive Integrated Moving Average (ARIMA) models, trained on 170 days of historical data. Linear Regression slightly outperformed others with a 1.40\% average absolute percentage difference in predictions. Our solution leverages a Cassandra NoSQL database, integrating heuristic optimization and shortage prediction to proactively manage blood resources. This scalable approach, designed for resource-constrained environments, considers factors such as proximity, blood type compatibility, inventory expiration, and rarity. Future developments will incorporate real-world data and additional variables to improve prediction accuracy and optimization performance.}

\onecolumn \maketitle \normalsize

\section{\uppercase{Introduction}}
\label{sec:introduction}
Blood reserve shortages represent a critical challenge for healthcare systems worldwide, particularly in low-income and disaster-prone areas. The World Health Organization (WHO) recommends that 1\% to 3\% of a country's population should donate blood to meet its basic healthcare needs~\cite{talie2020voluntary}. This guidance emphasizes the importance of regular and voluntary blood donations to ensure a consistent and adequate blood supply. However, numerous countries fall well below this threshold, creating tremendous strain on their healthcare systems~\cite{kralievits2015global,barnes2022status,roberts2019global}.
\begin{figure}[!h]
\vspace{-0.2cm}
\centering
\includegraphics[width=0.48\textwidth]{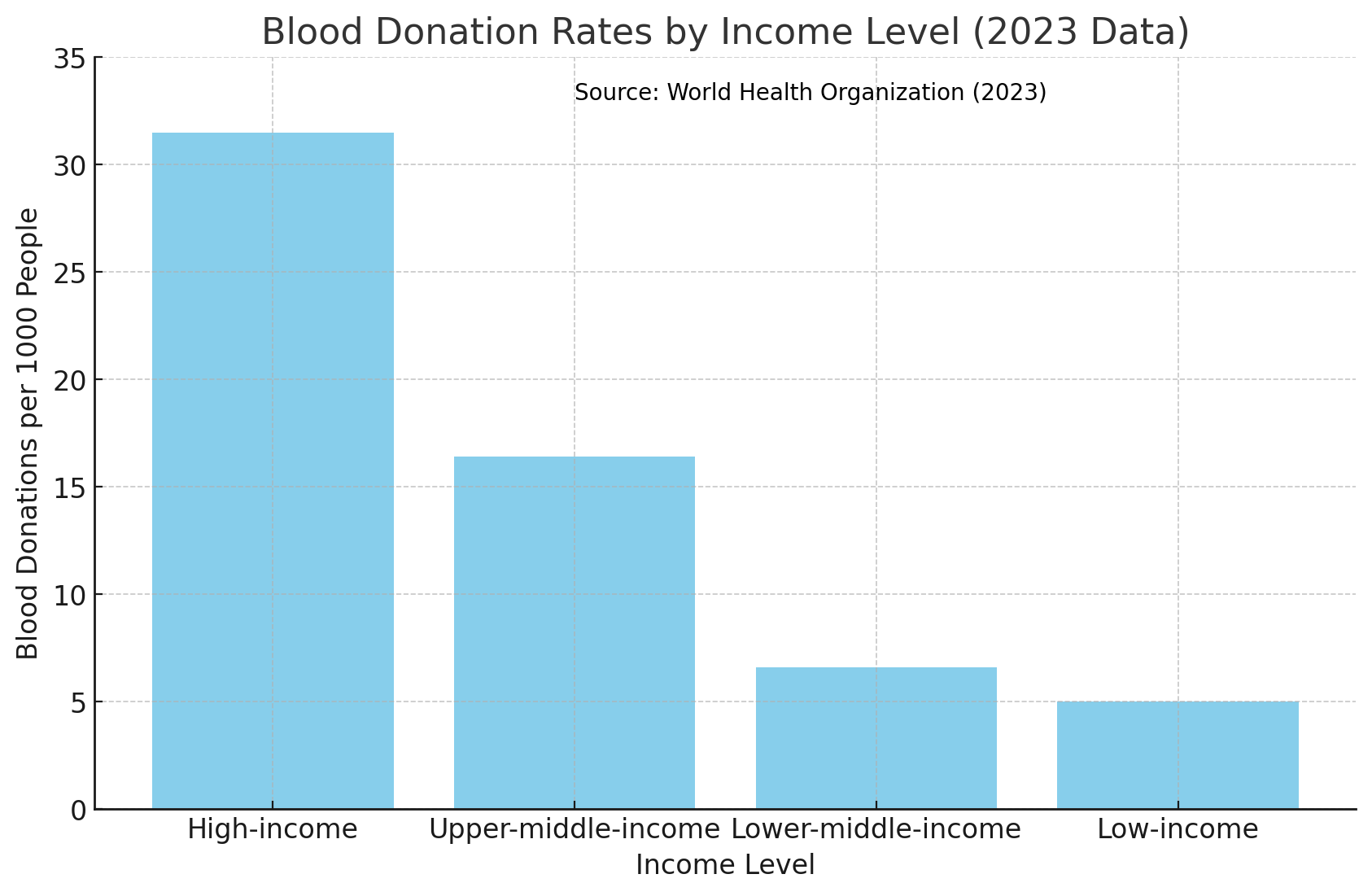}
\caption{Blood donation rates per 1000 people by country income level~\cite{who_blood_safety}.}
\label{fig:blood-donation-rates}
\end{figure}

Figure \ref{fig:blood-donation-rates} illustrates the significant disparity in blood donation rates across countries of different income levels. According to the WHO report from June 2023, these rates vary substantially based on a country's economic status~\cite{who_blood_safety}. The data, collected from samples of 1000 people, reveals a clear descending trend from high- to low-income countries. High-income countries lead with 31.5 donations, followed by upper-middle-income countries with 16.4 donations, lower-middle-income countries with 6.6 donations, and finally, low-income countries with the lowest rate of 5.0 donations.

These statistics underscore the disparities in blood donation rates across global economic divides. In low-income countries, the rate is as low as 5.0 donations per 1,000 people, which falls dramatically short of the World Health Organization's median recommendation of 10–20 donations per 1,000 people to meet essential healthcare needs. This gap of approximately 67\% below the recommended levels highlights the severe challenges these countries face in maintaining adequate blood supplies. Such shortages strain healthcare systems, especially in responding to medical emergencies and routine transfusion requirements, ultimately compromising patient care and increasing mortality risks~\cite{raykar2015blood}.


Another critical factor to consider is the heightened occurrence of blood shortages in areas affected by natural disasters or conflict, where the demand for blood can spike unpredictably, overwhelming already overburdened healthcare systems. For example, the COVID-19 pandemic caused significant disruptions to global blood donations due to lockdowns, fear of infection, and limited access to blood collection sites. This led to severe blood shortages in countries like the United States, resulting in postponed surgeries and delayed transfusions, which had a serious impact on patient care~\cite{riley2021public}. Similarly, during the Delta variant surge, the blood supply was critically low, necessitating rapid inventory management to ensure continuity of clinical care~\cite{petersen2023blood}. These crises highlight the need for robust, easily deployable systems that can optimize blood transactions and predict shortages in resource-constrained environments to mitigate the impacts of sudden demand surges~\cite{van2023managing}.


Moreover, in many low- and middle-income countries, blood donation systems heavily depend on family or replacement donors, with patients or their families often responsible for transferring blood between centers. For instance, in regions such as Africa and Latin America, families frequently coordinate blood donations and transport due to inadequate centralized systems \cite{who_blood_safety_2022}. Greece, where family donations are vital for conditions like thalassemia, also exemplifies how families play a crucial role in managing blood transfers \cite{ash_blood_services_2020}.

Given these challenges, there is a pressing need for innovative strategies to optimize blood supply management and accurately predict shortages, particularly in resource-limited areas. The WHO advocates for a transition to 100\% voluntary, unpaid donations as a critical step towards establishing sustainable blood supplies in these regions. However, achieving this objective requires more than policy reforms—it calls for the creation of scalable, reliable systems capable of functioning effectively even under high-pressure conditions and resource constraints~\cite{laermans2022impact}.

This paper introduces a novel approach that applies heuristic methods and machine learning techniques independently to optimize blood transactions and predict shortages in resource-constrained areas. Our system leverages the power of data-driven predictive modeling along with domain-specific heuristics to create a more efficient and effective blood supply management solution.

Key features of our proposed system include:

\begin{itemize}
    \item Heuristic approaches: These approaches include the utilization of blood compatibility matrices to maximize the utility of available blood supplies, the implementation of a rarity score for each blood group to prioritize scarce resources, consideration of blood product expiration dates to minimize wastage, and optimization of travel distances to improve logistics and reduce costs.
    
    \item Machine learning components: The machine learning components involve predictive models to forecast a center's request acceptance rate over a 10-day period and data-driven algorithms to optimize the direction of blood donations.
    
    \item Integration of heuristics and machine learning: A combined approach leverages the strengths of both methodologies, employing goal-oriented optimization to maximize the percentage of accepted blood requests.
\end{itemize}

This system is designed to tackle the significant disparities in blood donation and distribution, with a specific emphasis on low-income countries, conflict-affected areas, and regions vulnerable to natural disasters. Our solution aims to offer a scalable framework that can enhance blood supply management in these high-need regions, addressing the complex factors that affect blood donation, storage, and distribution.
\section{\uppercase{Background}}
\label{sec:background}
\subsection{Blood Type Distribution and Compatibility}

Understanding the distribution of blood types in a population and their compatibility is crucial for optimizing blood donation and distribution strategies. This section provides an overview of typical blood type distribution and compatibility.

\subsubsection{Blood Type Distribution}

The distribution of ABO blood types can vary across different populations. Table \ref{tab:blood-distribution} presents a typical distribution based on data from a large sample:

\begin{table}[h]
\centering
\caption{ABO blood types distribution.}
\label{tab:blood-distribution}
\begin{tabular}{|c|c|}
\hline
Blood Type & Abundance \\
\hline
A & 33.4\% \\
B & 6\% \\
O & 56.8\% \\
AB & 3.8\% \\
\hline
\end{tabular}
\end{table}

This distribution shows that Types A and O are the most common, followed closely by Type B, while Type AB is the least common. These percentages can inform predictions about blood supply and demand in a given population~\cite{sarhan2009distribution}.

\subsubsection{Blood Type Compatibility}

\begin{figure}[!h]
\vspace{-0.2cm}
\centering
\includegraphics[width=0.45\textwidth]{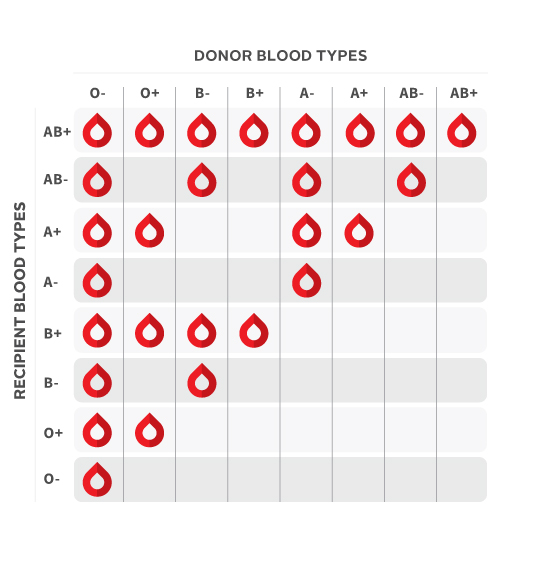}
\vspace{-0.7cm}
\caption{Blood type compatibility matrix~\cite{blood_compatibility_2024}.}
\label{fig:blood-compatibility}
\end{figure}

Blood type compatibility is a critical factor in transfusions and donation strategies. Figure \ref{fig:blood-compatibility} illustrates the compatibility matrix between different blood types.

The compatibility matrix illustrates the safe blood donation and transfusion pathways between different blood types. Key insights include the following:

\begin{itemize}
    \item Type O is the universal donor, able to give blood to all other types.
    \item Type AB is the universal recipient, able to receive blood from all other types.
    \item Types A and B can donate to their own type and to AB.
\end{itemize}

The blood type distribution data and compatibility matrix are integral to our methodology. We use this information to simulate realistic donor populations and blood transaction scenarios. Moreover, the compatibility matrix informs our heuristic search algorithm, enabling efficient matching of blood requests with available inventory. Additionally, we assign rarity scores to each blood group based on this data, which plays a key role in optimizing the allocation process.

\section{\uppercase{Related Work}}
\label{sec:related}

\subsection{Blood Bank Management Systems}

\begin{figure}[!h]
    \centering
    \includegraphics[width=1\linewidth]{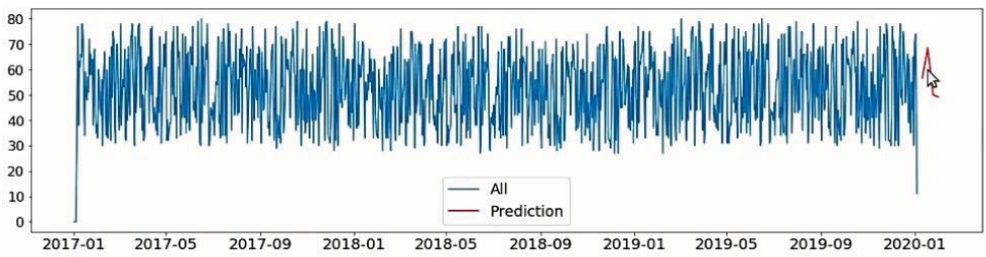}
    \caption{Blood demand prediction: actual values versus predicted values~\cite{sandaruwan2020towards}.}
    \label{fig:lstm_srilanka}
\end{figure}

Recent advancements in blood bank management have increasingly focused on leveraging emerging technologies to address operational inefficiencies. A significant contribution by ~\cite{sandaruwan2020towards} integrates machine learning, data clustering, and blockchain technologies. A key innovation is the application of Long Short-Term Memory (LSTM) networks for blood demand forecasting, which optimizes inventory management by improving demand prediction accuracy and operational efficiency. Figure \ref{fig:lstm_srilanka} illustrates the comparison between actual and predicted blood demand across all blood types.

Their system was validated using empirical data from the Blood Bank of Sri Lanka, with results indicating a strong potential for accurately forecasting future blood demand. The incorporation of blockchain technology enhances the system by introducing an additional layer of security, ensuring the integrity, traceability, and transparency of the blood supply chain.

Several critical points warrant consideration regarding the methodology and experimental design of this study: 1) the authors employed a highly imbalanced data split (98.2\% for training and 0.2\% for testing), which may raise concerns about the possibility of overfitting and could affect the reliability of the reported performance metrics; and 2) while the LSTM approach is promising, the study does not provide a comparison with other time series models, making it difficult to fully evaluate the relative effectiveness of this method in the context of blood demand forecasting~\cite{sandaruwan2020towards}.

Beyond Sandaruwan's work, ~\cite{ben2023smart} proposed a smart platform that uses machine learning and time series forecasting models to reduce blood shortages and waste by balancing blood collection and distribution. Their approach achieved an 11\% increase in collected blood volume and a 20\% reduction in inventory wastage compared to historical data. Moreover, \cite{shih2019comparison} compared machine learning algorithms and traditional time series methods like ARIMA for forecasting blood supply in Taiwan's blood services. Their results showed that time series methods, particularly seasonal ARIMA, outperformed machine learning models in predicting blood demand with better accuracy and lower error rates.

Recent studies also explore blockchain applications in blood bank management. \cite{wijayathilaka2020secured} developed the ``LifeShare" platform, incorporating blockchain technology for secure, transparent tracking of donors and blood supply, coupled with machine learning algorithms for blood demand forecasting. 

Despite these advancements, challenges remain. For example, \cite{farrington2023deep} applied deep reinforcement learning (DRL) to optimize platelet inventory management. Their results demonstrated DRL's effectiveness in reducing wastage while ensuring sufficient supply for hospitals, offering a promising avenue for future research in optimizing other blood components.

These studies provide a robust foundation for further research, particularly in combining machine learning models with blockchain technologies to enhance the security, accuracy, and efficiency of blood bank operations.

Our work builds upon this concept by:

\begin{itemize}
    \item Employing a more rigorous model evaluation and benchmarking methodology to enhance the reliability of results.
    \item Integrating heuristic optimization techniques that account for blood type compatibility, rarity scores, and logistical factors such as distance traveled, in order to improve the blood transfusion acceptance rate.
    \item Extending predictive capabilities to forecast a center's request acceptance rate over a 10-day horizon.
    \item Developing an open-source platform capable of adapting to real-time fluctuations in blood supply and demand, facilitating collaboration and seamless deployment.
\end{itemize}

These improvements aim to deliver a more comprehensive and practical solution to the challenges of blood bank management, especially in resource-limited environments.

\section{\uppercase{Methodology}}
\label{sec:methods}
\subsection{Database Schema}

To optimize data management and storage within our blood bank management system, we employed a Cassandra NoSQL database. The schema design was carefully structured to represent the key entities and relationships within the system, enabling efficient data retrieval and manipulation to support our simulations and predictive models.

\begin{figure}[!h]
    \centering
    \includegraphics[width=0.48\textwidth]{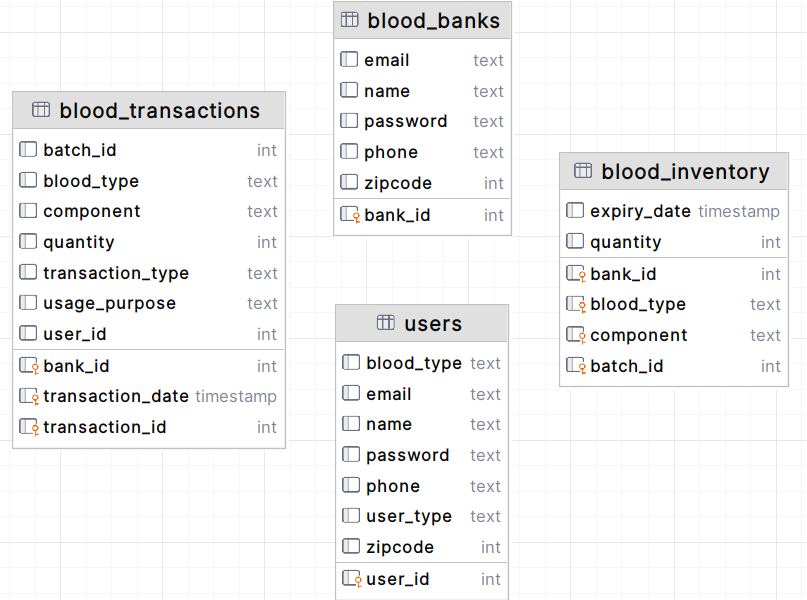}
    \caption{Cassandra database schema.}
    \label{fig:database-schema}
\end{figure}

Figure \ref{fig:database-schema} illustrates the database schema, which consists of four main tables:

\begin{enumerate}
    \item blood\_banks: Contains detailed information on individual blood banks, including contact information, geographic location, and relevant operational data.
    \item users: Stores comprehensive data for both donors and patients, including blood type, contact details, and associated medical information.
    \item blood\_inventory: Monitors the current inventory of blood at each bank, tracking quantities, expiration dates, and blood group availability.
    \item blood\_transactions: Logs all blood donation and request transactions, linking user data, blood bank operations, and inventory management for a comprehensive audit trail.
\end{enumerate}

Our study employed a two-pronged approach to optimize blood bank management: simulation-based optimization and shortage prediction modeling.

\subsection{Simulation-Based Optimization}

We developed three progressively advanced simulations to optimize blood bank operations:

\begin{enumerate}
    
    \item The initial simulation focused on the fundamental processes of blood requests and donations. Blood distribution was performed through random allocation, establishing baseline performance metrics to serve as a reference for evaluating more advanced optimization methods.
    
    \item The second simulation incorporated an optimization strategy based on geographic proximity, prioritizing blood banks nearest to the point of demand. A blood type compatibility matrix was integrated, and soon-to-expire blood units were prioritized to reduce wastage and improve resource utilization efficiency.
    
    \item The final simulation introduced rarity scores for various blood types to more effectively manage scarce resources. This simulation balanced proximity, expiration dates, and rarity in its allocation decisions, while also tracking expired units to identify areas for further optimization.
\end{enumerate}

Each simulation ran for a 30-day period, processing 40-50 transactions daily. We measured key performance indicators including acceptance ratio, total distance traveled, and number of expired resources.

\subsection{Shortage Prediction Modeling}

To forecast potential shortages, we developed and evaluated three time-series prediction models for comparative analysis:

\begin{enumerate}

    \item The first model, linear regression, was implemented as a baseline to capture linear trends in the acceptance ratios.
    
    \item The second model, Long Short-Term Memory (LSTM) networks, leveraged deep learning techniques to identify complex temporal patterns. It was trained on 170 days of historical data to generate predictive forecasts.
    
    \item Our final model, the Autoregressive Integrated Moving Average (ARIMA), applied traditional time-series forecasting techniques, effectively accounting for both trends and seasonality in the data.
    
\end{enumerate}

All models were trained to forecast the acceptance ratio for day 180, using data from the preceding 170 days. Model performance was assessed by calculating the mean absolute percentage error (MAPE) between the predicted and actual values.

\subsection{Performance Metrics}

To evaluate the effectiveness of our optimization strategies and predictive models, we employed the following metrics:
\begin{enumerate}
    \item Acceptance Ratio: Proportion of accepted blood requests.
    \item Total Distance Traveled: Sum of distances for all transactions.
    \item Marginal Performance (MP): Relative improvement in acceptance ratio.
    \item Average Absolute Percentage Difference: For assessing prediction accuracy.
\end{enumerate}

This comprehensive methodology enabled iterative improvements in blood bank operations through simulation, while also facilitating the development of accurate shortage prediction capabilities.

\section{\uppercase{Proposed Work}}
\label{sec:proposed}

\subsection{Dataset Synthesis and Description}

The sensitive nature of healthcare data, coupled with stringent privacy regulations governing blood donation records, presented considerable challenges in obtaining real-world data for this study. Health Information Privacy laws, such as the Health Insurance Portability and Accountability Act (HIPAA) in the United States, impose strict controls over the use and dissemination of patient health information, including blood donation records. Additionally, the cross-institutional structure of blood supply chains, involving multiple stakeholders, further complicates the process of comprehensive data collection due to logistical and legal constraints.

To address these limitations while maintaining the integrity of our analysis, we generated a synthetic dataset designed to replicate real-world blood bank operations. This approach enables us to explore a wide range of scenarios and rigorously test our algorithms, without compromising individual privacy or breaching data protection regulations. Our dataset comprises four main components:

\begin{enumerate}
    \item Blood Banks: A dataset comprising 20 simulated blood banks, each characterized by unique identifiers, names, geographic locations (represented by ZIP codes), and associated contact information.
    
    \item Users: A population of 1,000 individuals, categorized as donors, patients, or both. Each user has a unique identifier, demographic information, blood type, and ZIP codes.
    
    \item Blood Inventory: A comprehensive record of blood components (Red Blood Cells, Whole Blood, Platelets, and Plasma) available at each blood bank. Each inventory entry includes details such as blood type, quantity, expiration date, and a unique batch identifier.
    
    \item Blood Transactions: 4200 logs of donation events, linking donors to specific inventory batches. Each transaction includes information on the donor, recipient blood bank, blood component, quantity, and date of donation.
\end{enumerate}

To generate realistic user data, we employed the Faker library in Python~\cite{faker}, which offers a comprehensive set of methods for creating plausible synthetic data, such as names, addresses, phone numbers, and email addresses. Utilizing the Faker library allowed us to ensure that our synthetic dataset closely replicates real-world data while preserving anonymity and adhering to privacy standards.

A key aspect of our dataset is the distribution of blood types, which we based on global averages. Table \ref{tab:blood-type-distribution} shows the blood type distribution used in our simulation:

\begin{table}[!h]
\centering
\caption{Blood type distribution in synthetic dataset.}
\label{tab:blood-type-distribution}
\begin{tabular}{|c|c|}
\hline
Blood Type & Probability \\
\hline
O+ & 0.38 \\
A+ & 0.34 \\
B+ & 0.09 \\
O- & 0.07 \\
A- & 0.06 \\
AB+ & 0.03 \\
B- & 0.02 \\
AB- & 0.01 \\
\hline
\end{tabular}
\end{table}

Other key features of our dataset include the following: 1) Varied expiration dates for different blood components, accurately reflecting real-world storage constraints. 2) Simulated geographical distribution of blood banks and users, represented using ZIP codes to mimic real-world spatial dispersion. 3) Consistent linkage between inventory and donation transactions through unique batch identifiers, ensuring traceability and integrity of the data.

This synthetic dataset allows us to evaluate and validate our optimization algorithms under a range of scenarios, including rare blood type shortages, geographical constraints, and urgent inventory management challenges. While it does not fully capture the complexities of real-world blood bank operations, it provides a robust framework for testing and refining our blood supply chain optimization techniques, serving as a proof of concept for future development and more advanced applications.

The synthetic dataset, along with its accompanying schema and documentation, is made available for use by other researchers to facilitate further exploration and development in the field of blood supply chain optimization \cite{ieee_blood_donation_system}.

\subsection{System Design}

\begin{figure*}[t]
    \centering
    \includegraphics[width=0.9\textwidth]{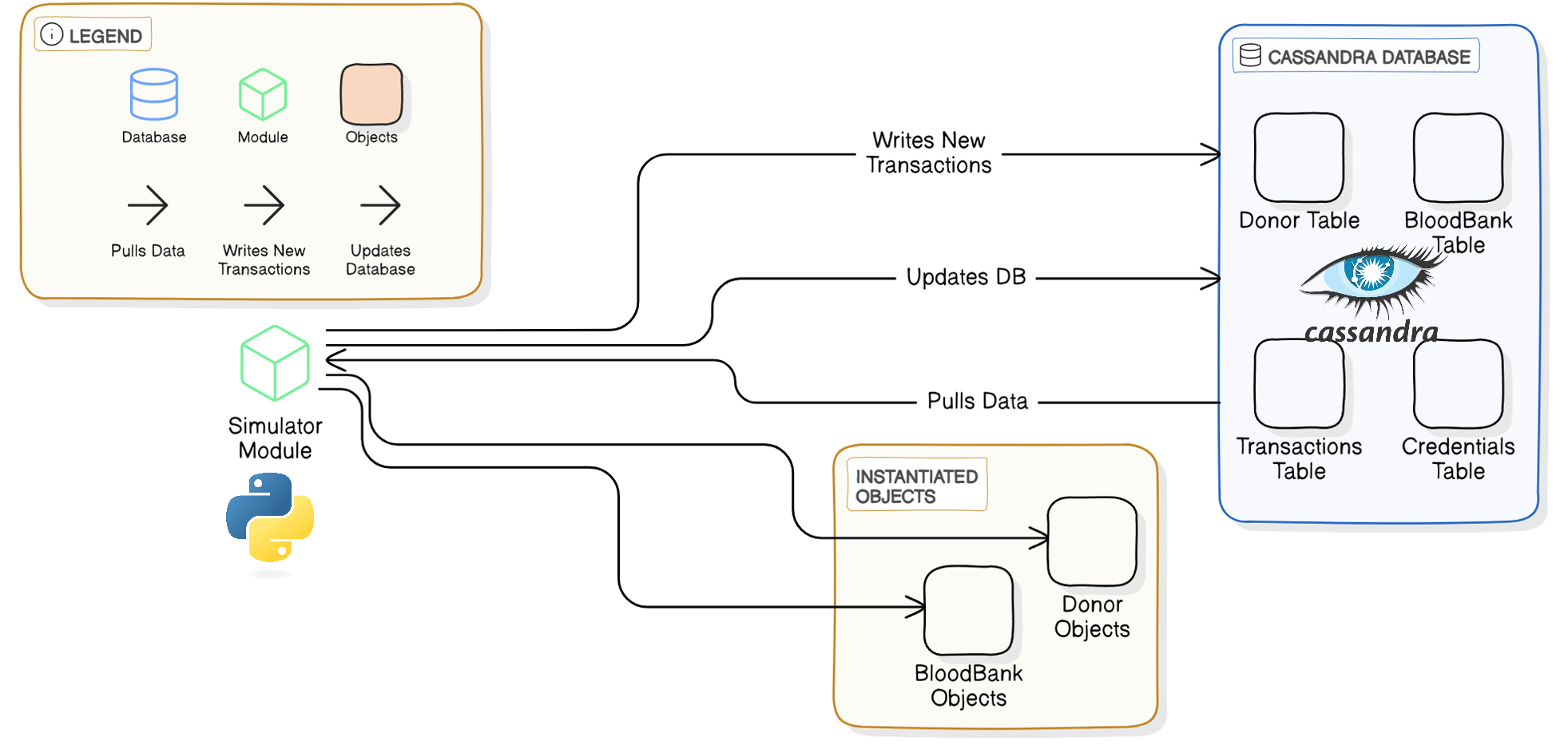}
    \caption{System architecture high-level view.}
    \label{fig:system-architecture}
\end{figure*}

Figure \ref{fig:system-architecture} illustrates the high-level architecture of our blood donation optimization system. At the core of our design is the Apache Cassandra NoSQL database~\cite{abramova2013nosql}, chosen for its ability to handle large-scale, distributed data with high availability and partition tolerance.

\subsubsection{CAP Theorem and Cassandra}

The Consistency, Availability and Partition Tolerance (CAP) theorem, a fundamental principle in distributed database systems, states that it is impossible for a distributed data store to simultaneously provide more than two out of the following three guarantees:

\begin{figure}[!h]
    \centering
    \includegraphics[width=0.6\linewidth]{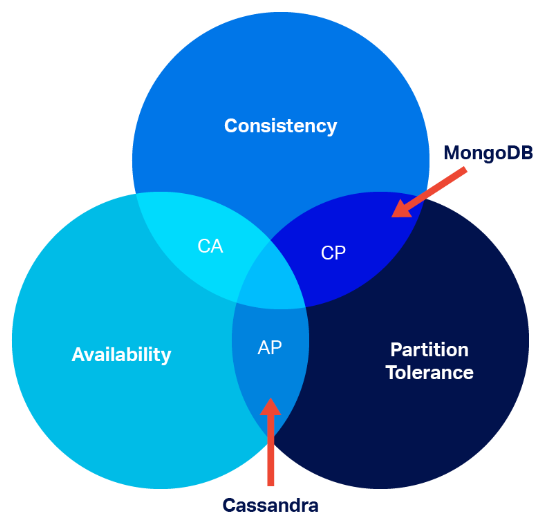}
    \caption{Visual representation of the CAP Theorem.}
    \label{fig:cap-theorem}
\end{figure}
As illustrated in Figure \ref{fig:cap-theorem}, the three guarantees are:

\begin{itemize}
    \item Consistency: Every read receives the most recent write or an error.
    \item Availability: Every request receives a response, without guarantee that it contains the most recent version of the information.
    \item Partition tolerance: The system continues to operate despite arbitrary partitioning due to network failures.
\end{itemize}

\subsubsection{Importance of AP in Blood Donation Systems}

The AP characteristics of Cassandra are vital to our system for several reasons. High availability is crucial, especially in emergencies, where the system must remain operational. Cassandra's multi-master architecture ensures continued functionality even if some nodes fail, unlike traditional SQL databases, which rely on a single master and are vulnerable to downtime in such cases~\cite{yedilkhan2023performance}. Scalability is another key factor, as blood donation networks span large regions, and Cassandra’s horizontal scaling allows seamless growth—something regular SQL databases struggle with due to their inherent vertical scaling limitations. While we trade strong consistency for eventual consistency, it suits most scenarios, and tunable consistency levels offer flexibility when needed. Cassandra’s partition tolerance also ensures the system stays operational during network issues. Furthermore, its native querying language, CQL, is powerful and efficient, providing a robust interface for managing complex data operations, far beyond the capabilities of traditional SQL for distributed data environments~\cite{perez2015modeling}.


    
    
    

\subsubsection{System Components}

As shown in Figure \ref{fig:system-architecture}, our system comprises the following key components:

\begin{itemize}
    \item Database: Stores data in four main tables: Donor, BloodBank, Transactions, and Credentials.
    
    \item Simulator Module: Acts as the central processing unit, pulling data from Cassandra, instantiating Donor and BloodBank objects, and simulating blood donation transactions.
    
    \item Data Flow: The simulator pulls data from Cassandra, processes it, writes new transactions, and updates the database, creating a continuous cycle of data management.
\end{itemize}

This architecture enables efficient data management, real-time processing of blood donation transactions, and seamless scalability to support the expansion of donor networks and blood banks. By utilizing Apache Cassandra as our database solution, we ensure that the system remains highly available and partition-tolerant—both of which are essential in the context of time-sensitive, geographically distributed blood donation systems. These capabilities are critical for maintaining uninterrupted operations and reliable access to blood inventory data, even in the presence of network disruptions or expanding system demands.

\subsection{Implementation Details}

\subsubsection{Simulation 1: Randomized Transactions}

Our first simulation model, termed ``Randomized Transactions," is designed to replicate the dynamic nature of blood donation and request processes within a network of blood banks. Implemented in Python, this simulation leverages the pandas library for efficient data manipulation and processing. The key components and processes of this simulation are as follows:

\begin{itemize}
    \item Our simulation uses two key data structures: the BloodBank class, which manages inventory, requests, and expired resources, and Pandas data frames to store and manipulate data for blood banks, users, inventory, and transactions, ensuring efficient data management and analysis.
    \item The simulation parameters span a 30-day period from January 1 to January 30, 2023, encompassing all eight major blood types (A+, A-, B+, B-, AB+, AB-, O+, O-) and four key blood components: Red Blood Cells (RBC), Platelets (PLAT), Plasma (PLAS), and Whole Blood (WB). Each blood component is assigned a specific expiration period: 42 days for RBC, 365 days for PLAS, 5 days for PLAT, and 35 days for WB. These expiration parameters closely align with real-world medical guidelines, ensuring the simulation accurately reflects the constraints of blood component storage and usage.~\cite{aubron2018platelet}.
    \item The simulation tracks two key performance metrics—acceptance ratio of blood requests and total distance traveled for donations and requests.

\end{itemize}
This randomized simulation serves as a baseline for analyzing the dynamics of blood supply and demand within a network of blood banks. It provides foundational insights into the variability and flow of donations and requests, offering a reference point for evaluating more advanced optimization techniques.

\subsubsection{Simulation 2: Heuristic Approach with Proximity and Expiration Prioritization}

Building on the foundation of Simulation 1, our second simulation model incorporates several heuristic optimizations to enhance blood allocation efficiency and minimize waste. The key new features introduced in this model are three heuristic elements to optimize blood allocation. First, it utilizes proximity-based selection, identifying the closest 15\% of blood centers for each request, thereby reducing transportation time and enhancing logistical efficiency. Second, the system incorporates expiration prioritization, ensuring that blood units nearing their expiration dates are allocated first, minimizing waste from expired products. Lastly, blood type compatibility is managed through a comprehensive compatibility matrix (refer to Figure \ref{fig:blood-compatibility}), implemented as a map data structure for efficient lookup, allowing for flexible and accurate blood type matching across the network.

\subsubsection{Simulation 3: Heuristic Approach with Rarity Scores}
\begin{algorithm}[!h]
 \caption{Simulation for blood bank management with expiry prioritization and rarity Scores.}
 \label{alg:algo3}
 \KwData{Blood banks, users, inventory data, blood type compatibility matrix, expiration rules, rarity scores}
 \KwResult{Updated transactions and inventory after simulation}

 Initialize simulation period, blood types, component expiration rules, and rarity scores\;
 Define blood type compatibility matrix\;
 \ForEach{day in the simulation period}{
    Remove expired blood based on expiration rules for each bank's inventory\;
    \While{daily operations not complete}{
        \eIf{Random event is a request}{
            Find 15\% closest blood banks to the user\;
            Check inventory using blood compatibility matrix and prioritize soon-to-expire resources\;
            \eIf{Blood bank can fulfill request (considering compatibility, expiration, and rarity)}{
                Fulfill request and update inventory\;
                Track distance, accepted requests, and selected blood type based on rarity scores\;
            }{
                Deny request and track failed attempts\;
            }
        }{
            Select a blood bank for donation\;
            Add donation to the bank’s inventory with an expiry date\;
            Track distance and donation data\;
        }
    }
 }
 Calculate performance metrics (acceptance ratio, total distance)\;
 Save updated transactions and inventory data\;
 
\end{algorithm}

Simulation 3 extends the functionality of Simulation 2 by incorporating a new heuristic: rarity scores for blood types. This enhancement further optimizes blood allocation by accounting for the relative scarcity of each blood type. The rarity score system, where lower scores represent rarer blood types, is implemented using a map data structure to facilitate efficient lookups and decision-making during allocation. Incorporating scarcity alongside other key factors enhances resource utilization, ensuring more effective allocation, particularly for rare blood types.

\begin{table}[!htb]
\centering
\caption{Blood type rarity scores.}
\label{tab:rarity-scores}
\begin{tabular}{|c|c|}
\hline
Blood type & Rarity score \\
\hline
O+ & 4 \\
A+ & 3 \\
O- & 3 \\
B+ & 2 \\
A- & 2 \\
B- & 1 \\
AB+ & 1 \\
AB- & 1 \\
\hline
\end{tabular}
\end{table}

The rarity scores are inversely correlated with the prevalence of each blood type in the general population. For example, O+ has the highest score due to its commonality, while AB- has the lowest score, reflecting its rarity. This scoring system enables the algorithm to prioritize the conservation of rarer blood types during allocation decisions, ensuring more efficient management of scarce resources.

This simulation retains the proximity-based selection, expiration prioritization, and blood type compatibility matrix from Simulation 2 (refer to Figure \ref{fig:blood-compatibility} for the compatibility matrix). The integration of rarity scores with these existing heuristics in Algorithm \ref{alg:algo3} creates a more comprehensive approach to blood bank management, potentially offering insights into strategies for managing blood inventories more effectively, especially for rare blood types.

\subsubsection{Shortage Prediction Techniques}

To further enhance our blood bank management system, we implemented a shortage prediction module designed to forecast the acceptance ratio of blood requests for each blood bank up to 10 days in advance. This predictive capability provides valuable insights, enabling proactive donation efforts and more efficient resource planning.

\begin{figure}[!h]
    \vspace{-0.2cm}
    \centering
    \includegraphics[width=\linewidth]{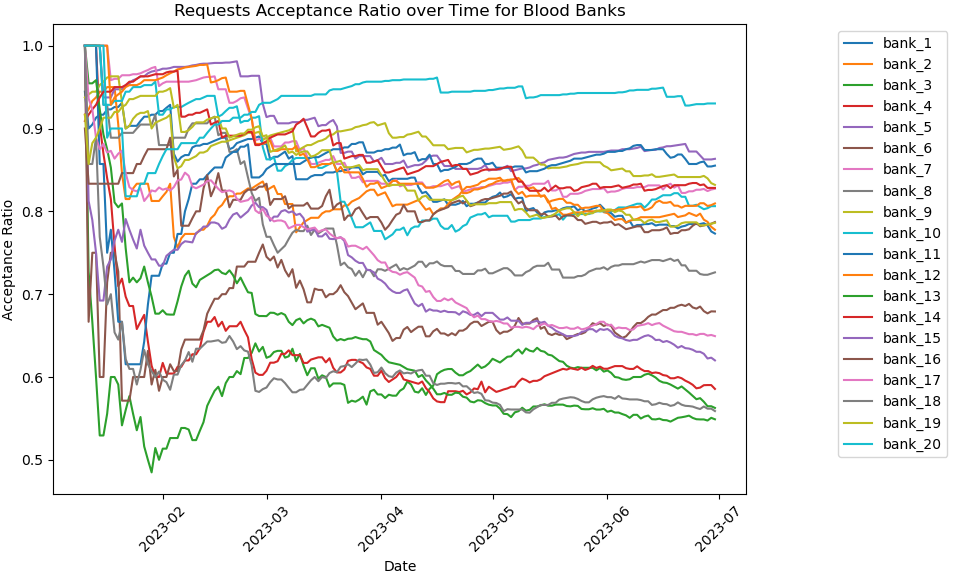}
    \caption{Acceptance ratio changes for blood banks over 180 days.}
    \label{fig:ratio-changes}
\end{figure}

We collected acceptance ratio data for each blood bank over a period of 180 days. This time series data forms the basis of our prediction models. Figure \ref{fig:ratio-changes} illustrates the variation in acceptance ratios across all blood banks over this period.

We implemented and compared three distinct time series prediction models, each offering unique strengths in processing time-dependent data:

\begin{enumerate}
    \item \textbf{Long Short-Term Memory (LSTM) Networks}: A type of recurrent neural network capable of learning long-term dependencies, well-suited for capturing complex patterns in time series data.
    
    \item \textbf{Linear Regression}: A simple yet effective approach for modeling linear trends in time series, serving as a baseline for comparison with more complex models.
    
    \item \textbf{Autoregressive Integrated Moving Average (ARIMA)}: Combines autoregression, differencing, and moving average components, effective for capturing various temporal structures in the data.
\end{enumerate}

Each model was trained on the 170-day historical data and tasked with predicting the acceptance ratio 10 days into the future for each blood bank.

\section{\uppercase{Results and Discussion}}
\label{sec:results}
\subsection{Database Query Samples}

To demonstrate the practical application of a NoSQL database within the blood donation management system, we present two sample queries written in Cassandra Query Language (CQL) along with their corresponding results.

\begin{figure}[!h]
    \centering
    \includegraphics[width=\linewidth]{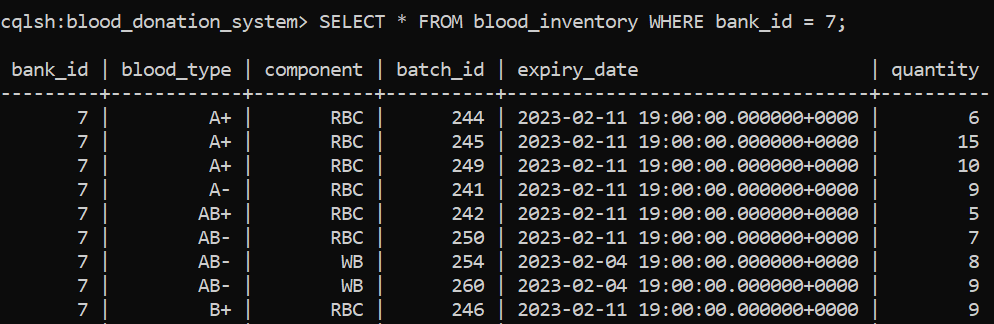}
    \caption{Query 1 to retrieve blood inventory for a specific bank.
    \label{fig:query-sample-1}
    }

\end{figure}

Figure \ref{fig:query-sample-1} illustrates a query that retrieves the complete blood inventory for a specific blood bank (bank ID = 7). This query efficiently accesses detailed inventory information, including blood types, components, batch IDs, expiration dates, and quantities. Such queries are essential for our optimization algorithms, enabling precise assessment of blood unit availability at a given location.

\begin{figure}[!h]
    
    \centering
    \includegraphics[width=\linewidth]{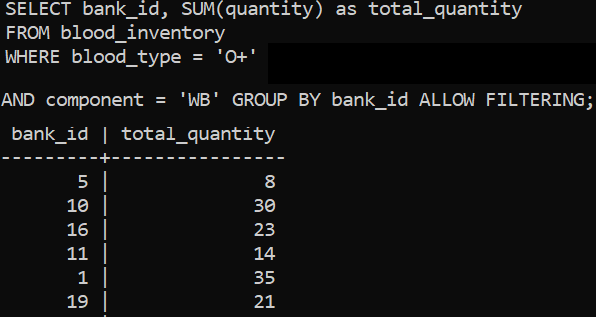}
    \caption{Query 2 to aggregate O+ whole blood quantities across banks.}
    \label{fig:query-sample-2}
\end{figure}

Figure \ref{fig:query-sample-2} presents a more complex query that aggregates the total quantity of O+ whole blood across multiple blood banks. This type of query is especially valuable for our shortage prediction models, enabling analysis of the distribution and availability of specific blood types and components across various locations.

These query examples demonstrate the flexibility and power of our database schema in supporting both detailed inventory management and high-level analytics. The ability to efficiently retrieve and aggregate data in this manner is fundamental to the performance of our optimization and prediction algorithms.

We conducted three simulations to evaluate the performance of our blood bank management system under different optimization strategies. This section presents the results and analyzes the improvements achieved with each iteration.

\subsection{Simulation Results}

Table \ref{tab:simulation-results} summarizes the key performance metrics for each simulation. We will analyze each of those.

\begin{table}[!h]
\caption{Performance metrics across the three simulations.}
\centering

\label{tab:simulation-results}
\resizebox{\columnwidth}{!}{%
\begin{tabular}{|l|c|c|c|}
\hline
Metric & Randomized & Heuristic 1 & Heuristic 2 \\
\hline
Total Accepted Requests & 553 & 679 & 674 \\
Total Denied Requests & 147 & 123 & 84 \\
Overall Acceptance Ratio & 0.79 & 0.85 & 0.89 \\
Total Units Traveled & 2,496,171 & 1,573,463 & 1,551,953 \\
\hline
\end{tabular}%
}
\end{table}

\subsubsection{Acceptance Ratio Analysis}

The acceptance ratio demonstrated consistent improvement across the three simulations, indicating the effectiveness of the progressively refined heuristics.

In the first comparison, moving from Simulation 1 to Simulation 2, we observed an increase from 79\% to 85\%, a change that was statistically significant with a \( z \)-score of 2.79 and a \( p \)-value of 0.0026. This marked an improvement in optimizing blood allocation.

The second comparison, between Simulation 2 and Simulation 3, showed further gains, with the acceptance ratio rising from 85\% to 89\%. This improvement, although smaller, remained statistically significant, with a \( z \)-score of 2.12 and a \( p \)-value of 0.0170.

Finally, the cumulative effect from Simulation 1 to Simulation 3 was even more striking, with the acceptance ratio increasing from 79\% to 89\%. This overall improvement was highly significant, with a \( z \)-score of 4.90 and a \( p \)-value of less than 0.00001, underscoring the substantial impact of incorporating additional heuristic elements.

\subsubsection{Distance Optimization}
The total distance traveled decreased significantly:
\begin{itemize}
    \item Sim 1 to Sim 2: Reduced by 922,708 (36.96\% reduction)
    \item Sim 2 to Sim 3: Further reduced by 21,510 (1.37\% reduction)
    \item Overall reduction (Sim 1 to Sim 3): 944,218 (37.83\% reduction)
\end{itemize}

The distance reduction analysis reveals that the first heuristic introduced in Simulation 2 was the most effective in minimizing the distance traveled, accounting for nearly all of the overall gains (36.96\%). The second heuristic, while contributing sightly, resulted in diminishing returns with only 1.37\% additional reduction. Therefore, most of the optimization occurred during the first transition, indicating that further improvements in distance reduction may require more advanced or different heuritics beyong those applied in Simulation 3. 

\subsubsection{Marginal Performance (MP) Analysis}

To quantify the incremental benefits of each simulation, we calculate the Marginal Performance (MP) using equation \ref{eq1}:

\begin{equation}\label{eq1}
\Delta MP = \frac{\text{Acceptance rate} - \text{Baseline acceptance rate}}{1 - \text{Baseline acceptance rate}}
\end{equation}

Table \ref{tab:mp-analysis} presents the results of this analysis:

\begin{table}[H]
\caption{Marginal performance analysis.}
\centering
\label{tab:mp-analysis}
\resizebox{\columnwidth}{!}{%
\begin{tabular}{|l|c|c|}
\hline
Transition & $\Delta$MP (Accept ratio) & $\Delta$MP (Distance reduced) \\
\hline
S1 to S2 & 28.8\% & 37\% \\
S2 to S3 & 26.7\% & 1.4\% \\
S1 to S3 & 47.6\% & 37.8\% \\
\hline
\end{tabular}%
}
\end{table}

The MP analysis demonstrates that the largest improvements in both acceptance ratio and distance traveled occur during the transition from S1 to S2. While acceptance ratio continues to improve in the transition S2 to S3, the gains in distance reduction are minimal, suggesting diminishing returns in this aspect. Overall, the analysis shows that applying heuristic methods substantially improves performance, with nearly 48\% improvement in acceptance rate and a 38\% reduction in total distance traveled when comparing the final simulation to the baseline.

\subsection{Shortage Prediction Results}

We evaluated three predictive models—Long Short-Term Memory (LSTM), Linear Regression, and Autoregressive Integrated Moving Average (ARIMA)—for forecasting the acceptance ratio of blood banks. Each model was trained on 170 days of historical data and tasked with predicting the acceptance ratio for day 180. The results of each model's performance across 20 blood banks are presented below.


Table \ref{tab:model-comparison} summarizes the performance of each model:
\begin{table}[h!]
\centering
\caption{Model performance comparison.}
\label{tab:model-comparison}
\resizebox{\columnwidth}{!}{%
\begin{tabular}{|l|c|}
\hline
Model & Mean percent difference \\
\hline
LSTM & 1.48\% \\
Linear Regression & 1.40\% \\
ARIMA & 1.83\% \\
\hline
\end{tabular}%
}
\end{table}

Based on the mean percent difference, both the LSTM and Linear Regression models performed similarly well, with LSTM having a 1.48\% difference and Linear Regression slightly better at 1.40\%. ARIMA, while still performing reasonably, had a higher mean percent difference of 1.83\%. Overall, the results suggest that Linear Regression and LSTM are the most accurate models for predicting the acceptance ratio, with Linear Regression showing a slight edge in performance over the other two models.


Tables \ref{tab:lstm-results}, \ref{tab:linear-regression-results}, and \ref{tab:arima-results} show the detailed results for each model. Due to space constraints, we present results for a subset of blood banks.
\begin{table}[!h]
\centering
\caption{LSTM model results subset.}
\label{tab:lstm-results}
\small
\resizebox{\columnwidth}{!}{%
\begin{tabular}{|c|c|c|c|}
\hline
Bank ID & Predicted & Actual & Percent difference \\
\hline
1 & 0.7816 & 0.7757 & 0.7648\% \\
5 & 0.8771 & 0.8626 & 1.6785\% \\
10 & 0.8184 & 0.8063 & 1.5101\% \\
15 & 0.6610 & 0.6233 & 6.0379\% \\
20 & 0.9465 & 0.9302 & 1.7483\% \\
\hline
\end{tabular}%
}
\end{table}

\begin{table}[!h]
\centering
\caption{Linear regression model results subset.}
\label{tab:linear-regression-results}
\small
\resizebox{\columnwidth}{!}{%
\begin{tabular}{|c|c|c|c|}
\hline
Bank ID & Predicted & Actual & Percent difference \\
\hline
1 & 0.7807 & 0.7731 & 0.9736\% \\
5 & 0.8801 & 0.8634 & 1.9322\% \\
10 & 0.8205 & 0.8063 & 1.7704\% \\
15 & 0.6349 & 0.6201 & 2.3944\% \\
20 & 0.9390 & 0.9302 & 0.9380\% \\
\hline
\end{tabular}
}
\end{table}

\begin{table}[!h]
\centering
\caption{ARIMA model results subset.}
\label{tab:arima-results}
\small
\resizebox{\columnwidth}{!}{%
\begin{tabular}{|c|c|c|c|}
\hline
Bank ID & Predicted & Actual & Percent difference \\
\hline
1 & 0.7799 & 0.7731 & 0.8683\% \\
5 & 0.8801 & 0.8634 & 1.9407\% \\
10 & 0.8219 & 0.8063 & 1.9428\% \\
15 & 0.6535 & 0.6201 & 5.3974\% \\
20 & 0.9380 & 0.9302 & 0.8306\% \\
\hline
\end{tabular}%
}
\end{table}

All three models demonstrate strong performance, with average absolute percentage differences ranging from 1.40\% to 1.83\%. Among them, Linear Regression ranked the highest with 1.40\% difference, followed by LSTM at 1.48\%, while ARIMA had the highest difference at 1.83\%.

Despite their overall accuracy, all models struggled with specific outliers, such as Bank 15, indicating that certain blood banks may have more stochastic acceptance ratios. The findings suggest that simpler models like Linear Regression provide a good balance between accuracy and computational efficiency for this task. The results demonstrate the feasibility of using predictive models to forecast acceptance ratios, aiding resource allocation and shortage prevention in blood banks.\\








\vspace{-0.7cm}
\section{\uppercase{Conclusions and Future Work}}
\label{sec:conclusion}

In conclusion, our approach demonstrates significant improvements in blood supply management through the introduction of additional heuristic matching criteria such as proximity, blood compatibility matrix, rarity scores, and expiration dates. These enhancements led to a 38\% reduction in the total distance traveled, which is crucial in emergencies where timely delivery is essential. Furthermore, our multi-level heuristic matching strategy resulted in a 48\% improvement in matching performance $\Delta$MP, highlighting the potential for more efficient resource allocation.

Predicting blood transfusion acceptance rates over a 10-day window also played a vital role in optimizing blood supply usage. Our evaluation shows that while linear regression, LSTM, and ARIMA models perform similarly in terms of accuracy, however linear regression stands out as a simpler statistical predictive approach, offering comparable results with lower computational complexity.

For future work, we intend to explore the development of mobile and web applications designed to maximize user engagement, focusing on both donor participation and improving the user experience for patients, especially those requiring frequent transfusions~\cite{li2023mobile}. Additionally, we aim to create an integrated model that combines shortage prediction with heuristic matching. This combined approach will be thoroughly assessed to determine its impact on acceptance rates, comparing it to our current system to further enhance efficiency and effectiveness in blood supply management.

\vspace{-0.5cm}
\bibliographystyle{apalike}
{\small
\bibliography{example}}

\end{document}